\documentclass[conference]{IEEEtran}
\IEEEoverridecommandlockouts
\usepackage{cite}
\usepackage{amsmath,amssymb,amsfonts}
\usepackage{algorithmic}
\usepackage{graphicx}
\usepackage{textcomp,xcolor}
\usepackage{ulem} 
\usepackage{hyperref} 
\usepackage{svg}

\def\BibTeX{{\rm B\kern-.05em{\sc i\kern-.025em b}\kern-.08em
    T\kern-.1667em\lower.7ex\hbox{E}\kern-.125emX}}

                \newenvironment{where}{\noindent{}where\begin{itemize}}{\end{itemize}}
                
\newcommand{\vect}[1]{\boldsymbol{#1}}  
\begin{document}

\title{Data-Driven Supervision of a Thermal-Hydraulic Process Towards a Physics-Based Digital Twin \\
}

\author{\IEEEauthorblockN{Osimone IMHOGIEMHE}
\IEEEauthorblockA{\textit{Nantes Univ., Centrale Nantes} \\
\textit{LS2N, CNRS UMR 6004}\\
F-44000 Nantes, France \\
osimone.imhogiemhe@ls2n.fr}
\and
\IEEEauthorblockN{Yoann JUS}
\IEEEauthorblockA{\textit{Fluid \& Sealing Technologies} \\
\textit{CETIM}\\
Nantes, France \\
yoann.jus@cetim.fr}
\and
\IEEEauthorblockN{Hubert LEJEUNE}
\IEEEauthorblockA{\textit{Fluid \& Sealing Technologies} \\
\textit{CETIM}\\
Nantes, France \\
hubert.lejeune@cetim.fr}
\and

\IEEEauthorblockN{Saïd MOUSSAOUI}
\IEEEauthorblockA{\textit{Nantes Univ., Centrale Nantes} \\
\textit{LS2N, CNRS UMR 6004}\\
F-44000 Nantes, France \\
said.moussaoui@ec-nantes.fr}
}

\maketitle

\begin{abstract}
The real-time supervision of production processes is a common challenge across several industries. It targets process component monitoring and its predictive maintenance in order to ensure safety, uninterrupted production and maintain high efficiency level. The rise of advanced tools for the simulation of physical systems in addition to data-driven machine learning models offers the possibility to design numerical tools dedicated to efficient system monitoring. In that respect, the digital twin concept presents an adequate framework that proffers solution to these challenges. The main purpose of this paper is to develop such a digital twin dedicated to fault detection and diagnosis in the context of a thermal-hydraulic process supervision. Based on a numerical simulation of the system, in addition to machine learning methods, we propose different modules dedicated to process parameter change detection and their on-line estimation. The proposed fault detection and diagnosis algorithm is validated on a specific test scenario, with single one-off parameter change occurrences in the system. The numerical results show good accuracy in terms of parameter variation localization and the update of their values.    

\end{abstract}


\section{Introduction}
Thermal-hydraulic systems play an important role across different industries, relying on the interaction between thermal (heat transfer) and hydraulic (fluid flow) processes. These systems involve the transfer of heat and fluid flow, between components such as boilers, heat exchangers, pumps, and pipelines~\cite{b1}. These systems are fundamental in sectors such as power generation, manufacturing, chemical processing, HVAC (heating, ventilation, and air conditioning) and aerospace~\cite{b2}. Some of the problems faced include degradation, fouling and corrosion, which require effective maintenance policies to ensure system reliability and efficiency. The integration of sensors and simulation tools allows real-time control and modeling tools respectively. However, the acquired data requires specific expertise in addition to choosing accurate fault detection and diagnosis (FDD) algorithms to ensure the reliability, safety and efficiency of these systems~\cite{b3}.

\par Fault detection and diagnosis, a subsection of control engineering, is concerned with keeping track of a system, determining when a fault has occurred and locating the fault. Depending on the industry, maintenance accounts for a large amount of manufacturing costs, therefore a lot of research goes into fault diagnosis to reduce costs \cite{b5}. The FDD process is one of the critical components in the maintenance and operation of thermal-hydraulic systems. An effective FDD ensures the reliability, safety and efficiency of these systems. The FDD involves determining what a fault is, in order to determine the occurrence of these abnormal conditions. In the event of a detected fault, fault diagnosis determines the source of the problem, allowing for appropriate corrective actions. Some common FDD methods include statistical, model-based and data-based methods \cite{b6}. Data-based methods utilize machine learning to analyze fault data and determine causes \cite{b3}, while some hybrid methods leverage the strengths of all methods. It therefore requires large amounts of quality data, high computational capacity, real-time processing and advanced machine learning algorithms. 

\par The digital twin (DT) concept, proposed by Grieves in 2003 \cite{b7}, has become a pivotal innovation in various industrial domains. It is mainly based on an adaptive virtual representation of the physical system. However, the concepts and capabilities of Digital Twins (DTs) are not clearly defined and can sometimes be challenging to understand. This ambiguity arises because DTs can be utilized for various tasks across different life-cycle phases and industrial domains ~\cite{b4}~\cite{b15}. This also includes implementations for fault detection and diagnosis \cite{b21}, especially in hydraulic systems \cite{b22,b16,b17}. 

\par Consequently, different interpretations of the DT concept exist, driven by specific use cases even if several initiatives emerge to develop such standardized approach as the Industrial Digital Twin Association (IDTA), the Digital Twin Consortium and different standards: IEC 63278-1:2023 for example\cite{b17b}.
This results in the issue that concrete implementations of DTs pursue specific objectives without adhering to any standardized architectural template~\cite{b18}. Digital twin models can be classified based on the approach to system modeling and to numerical simulation. Among these are data-driven, physics-based and hybrid models which offer unique advantages and cater to different requirements \cite{b12,b13,b14}. Data-driven models leverage vast amounts of data generated from sensors, IoT devices, and other sources to build predictive and descriptive models. Physics-based models use physical laws to simulate the behavior of systems. Hybrid models offer a balanced approach, combining the adaptability of data-driven models with the accuracy of physics-based models.


\par This paper delves into the design of the digital twin of a thermal-hydraulic process system with a particular focus on fault detection and diagnosis. It particularly intends to perform the parameter estimation using machine learning algorithms trained with data generated by a numerical simulation of a physical model of the system. Section 2 presents the general framework of the study. On that foundation, Section 3 introduces the fault detection and diagnosis methodology. Section 4 highlights the results in order to validate our methodology and Section 5 provides conclusion and perspectives.

\section{General Framework}
\subsection{Digital Twin}

To address fault detection and diagnosis in the DT framework, the following criteria are assumed, using the guidelines proposed by Alliance Industrie du Futur (AIF) \cite{b21b}: 
\begin{itemize}
    \item The DT is an organized set of models representing a real-world entity, developed in response to specific problems and uses.
    \item The DT is updated in relation to the properties of its physical twin, with a frequency and precision adapted to its issues and uses.
    \item The DT is equipped with advanced operating tools enabling it to understand, analyze, forecast, optimize operation and control of the real entity.
\end{itemize}
The essentials of every digital twin framework are the physical twin, the virtual twin and the two-way data connectivity. Other unconsolidated components include the user interface, cloud connectivity, and machine learning for analytics and simulation.\\
\subsection{The Physical Process}
The physical asset 
is a closed water hydraulic loop equipped with three actuators (pump speed, opening of the control valve and opening of a cooling on/off valve) to regulate the pressure, flow and temperature in this loop. The pump controls the working fluid through the circuit. A control valve regulates the fluid flow in parallel to a heat exchanger, which regulates the temperature. The loop is pressurized by a tank connected to the site’s compressed air circuit. In order to generate some defects artificially, other control devices are installed on the physical twin: heating collars, one manual valve placed at pump suction (to generate cavitation) and one placed upstream of the exchanger (to degrade its performance). Finally, an air pressure regulator allows perturbation of the pressure level of the whole loop.

\begin{figure}[htbp]
\centerline{\includegraphics[width=0.5\textwidth]{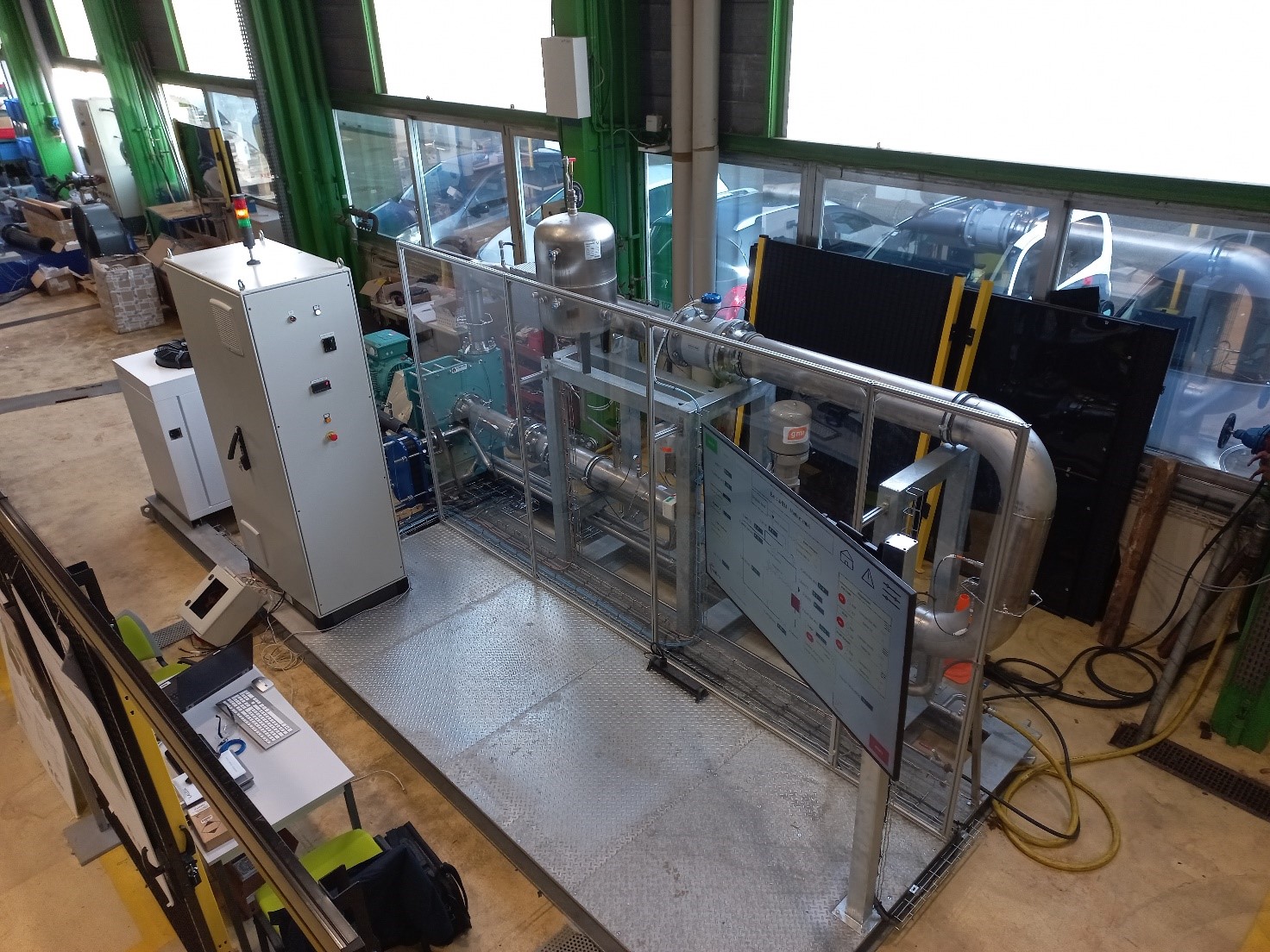}}
\caption{Photo of the closed hydraulic loop}
\label{fig:Photo_JNEM}
\end{figure}

\subsection{The Virtual Twin}
The virtual twin in the DT concept is more than just the 3-dimensional modeling of the physical twin, though a crucial part of the overall concept. It is required to have multi-parameter simulation capabilities. In Section 1, some of these simulation approaches were discussed. This section discusses the hybrid simulation model used for the DT design. The hydraulic system is divided into these major parts: actuators, process, control, and sensors. There is therefore a possibility of faults in each section of the system. The scope of this project focuses on component failures through a change in process conditions. Such component faults can be abrupt, gradual, and intermittent; each of these are considered in the project but results for abrupt fault simulations will be discussed in Section 4. The abrupt fault best models all three cases because until the threshold is exceeded, the current algorithm cannot detect a fault. In order to apply the digital twin concept in this project, the physics-based model is chosen as it provides more advantages with the ability to represent the system with mathematical equations. In so doing, a database can be created for machine learning faster than creating one with the physical bench. In order to further optimize on time, numerical simulations are preferred to analytical calculations. Computational fluid dynamics (CFD) and other advanced simulation tools are increasingly used to model and analyze thermal-hydraulic systems. 

\begin{figure}[htbp]
\centerline{\includegraphics[width=0.5\textwidth]{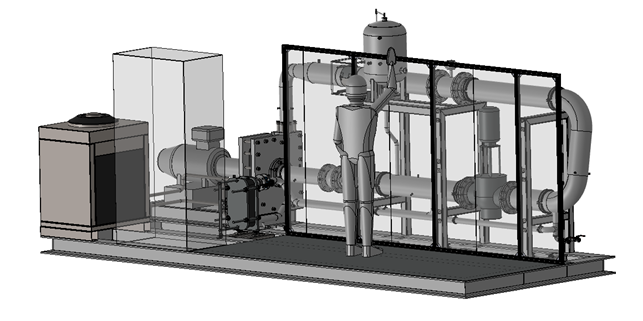}}
\caption{The CAD of the closed hydraulic loop}
\label{fig:CAO_JNEM}
\end{figure}

A 1D fluid simulation model is a physical model built using fluid mechanics equations related to the thermal-hydraulic process, simplified by the assumption that the characteristics of the fluid are homogeneous in each section of the flow. This type of modeling enables very rapid predictive calculations to be made, which could then be integrated into a control system in real-time or near-real-time. The 1D model provides a predictive digital model of the loop in terms of operating conditions. In order to enhance the representation and industrial transfer potential of the demonstrator, it was decided to give priority to using commercial software suitable for fluid network modelling to produce this model rather than developing a specific tool for the loop. Simcenter Flomaster (version 2401)~\cite{b20} was chosen for its ability to model complex fluid systems.

\begin{figure}[htbp]
\centerline{\includegraphics[width=0.5\textwidth]{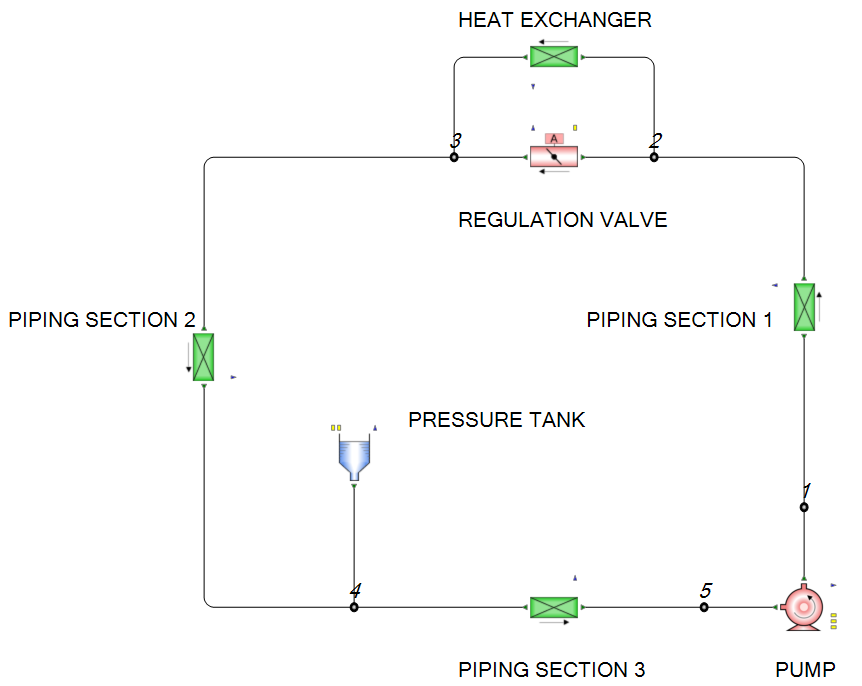}}
\caption{The 1D model of the closed hydraulic loop on Simcenter Flomaster software}
\label{fig:1D_JNEM}
\end{figure}

The modeling strategy consists of simplifying the model to limit the number of components and the number of model parameters. To this end, the various piping elements, exchangers and manual valves are represented by head loss components. Circuit pressure and flow are defined as the intersection of the pump curve (total head vs. flow) with the network curve (pressure drop vs. flow). Piping details are therefore not required for this approach. Only the head loss information for the various components is retained, which has the advantage of enabling rapid pressure and flow calculations. The model constructed is shown in Figure \ref{fig:1D_JNEM}.

\subsection{Input-Output Parameters}
The two actuators in this model are the control valve and the pump. The connection nodes between the different sections respect the location of the static pressure taps in the physical loop. Only pressure and flow are taken into account, neglecting the thermal parameters for this preliminary work. It is therefore a stationary incompressible isothermal 1D model of the loop. This 1D model has the following components: a) Pump, b) Regulation Valve, c) Heat Exchanger,  d) Tank, e) Piping Sections 1, 2 and 3.

The pump, working at a certain percentage of performance speed ($u_1$), and the flow in the system also controlled by the percentage opening of the regulation valve ($u_2$), which are the two input variables, are defined as the control vector represented by $U$. 

The flow from the pump passes through piping section 1 to the heat exchanger and the regulation valve then to piping section 2, the storage tank and the system is closed with piping section 3. There are five nodes in this model where hydraulic calculations are gotten showing the properties of the system, enabling a monitoring system; these nodes are: a) Output pressure of pump "P1" (bar), b) Input pressure of pump "P2" (bar), c) Input pressure of heat exchanger "P3" (bar), d) Output pressure of heat exchanger "P4" (bar), e) Hydraulic loop overall flow "FL" (m$^3$/h). These 5 variables are defined as the process (output) vector represented by $Y$.

The nominal process vector at any given control (input) vector is given by the nominal operating values of the working parts of the hydraulic loop. The parameters of the hydraulic loop in our case study are as follows: a) Head loss of piping section 1, b) Head loss of piping section 3, c)  Head loss of heat exchanger, d) Tank pressure (bar), e) Pump-rated head (m), f) Pump-rated flow (m$^3$/h). These parameters are called the ”component vector” and represented by $\theta$. The nominal values are reported in Table \ref{tab:parvalues}. 

\begin{table}[htbp]
\caption{\label{tab:parvalues} Nominal Values of the Component Vector}
\begin{center}
\begin{tabular}{|c|c|c|}
    \hline
    \textbf{Parameter} & \textbf{Description} & \textbf{Value} \\ [0.5ex]
    \hline
       $\theta_1$ & Head loss of piping section 1 "Loss1" (-) &  4.5\\
       $\theta_2$ & Head loss of piping section 3 "Loss3" (-) &  1.17\\
       $\theta_3$ & Head loss of heat exchanger "Lossx" (-) &  10.35\\
       $\theta_4$ & Tank pressure "P$\_$Tank" (bar)&  3\\
       $\theta_5$ & Pump rated head "HMT"(m) &  229\\
       $\theta_6$ & Pump rated flow "Debit"($m^3/h$) &  15.3\\
        \hline
\end{tabular}
\end{center}
\end{table}

\section{Fault Detection and Diagnosis Approach}
There are four questions that arise from any fault in the system. First, how can a change in the internal parameters of the component vector be accurately detected? Second, how can the changed parameter(s) be localized? Third, how can the new value be estimated? Finally, how can this whole process be validated?

\subsection{Monitoring principle}
\begin{figure}[htbp]
\centerline{\includegraphics[width=0.5\textwidth]{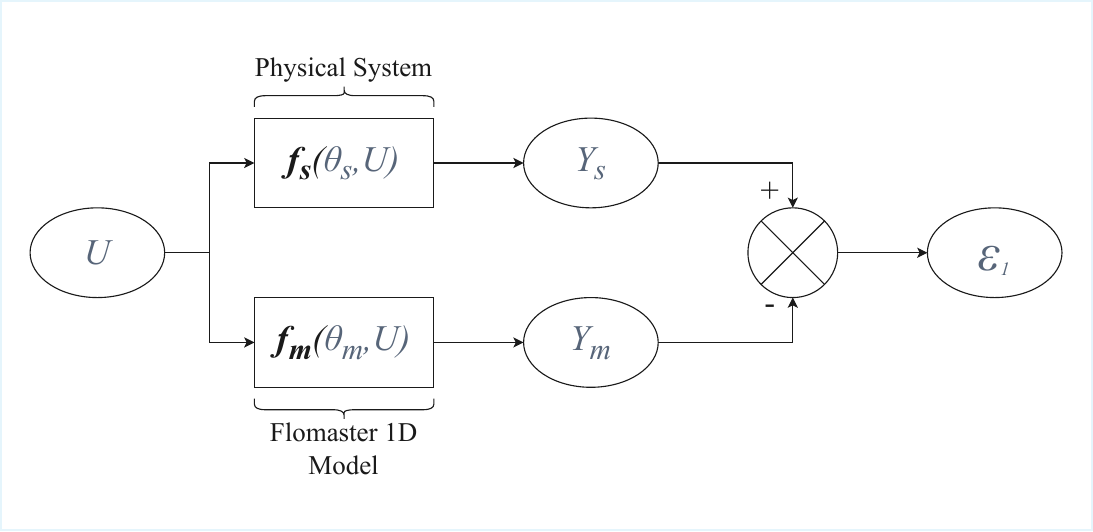}}
\caption{The simplified monitoring principle using the DT}
\label{fig:DT_JNEM}
\end{figure}

This component vector cannot be directly measured, but the initial set value is known at the design of the system. When running the system in any particular combination of control vectors, while the vector process $Y_s$ from the physical setup is the same as the vector process $Y_m$ from the Flomaster 1D model, it can be concluded that the component vector $\theta$ is known. The values, in such a case, are therefore the same as configured inside the Flomaster software. In addition to noisy sensor data or approximation error, any difference between $Y_s$ and $Y_m$ at the same $U$ as in Figure \ref{fig:DT_JNEM} means that the component vector of the physical setup does not match what is in the Flomaster 1D model.

In order to detect the changed parameter in the component vector and then estimate the new value(s), an algorithm is introduced that is triggered by a difference between $Y_s$ and $Y_m$. The threshold is manually tuned and partly based on the precision of the sensors, sensitivity analysis carried out on each component testing the variation of each component vector and their effect on the process vector.
\begin{align}\label{eqn:detection}
\varepsilon^k &= |y_{s}^{k} - y_{m}^{k}|\\
\varepsilon^k &< \varepsilon^k_d
\end{align}

\begin{where}
  \item $k$ is the index of the vector from 1 to 5 for the process vector.
  \item $\varepsilon^k$ is the kth error value in the error vector.
  \item $y_{s}^{k}$ is the kth value of the measured system process vector.
  \item $y_{m}^{k}$ is the kth value of the simulated 1D model process vector.
  \item $\varepsilon^k_d$ is the kth value of the set detection threshold.
\end{where}

\subsection{Fault Detection}
The detection of the fault is based on a shift in the measurement vector. A deviation of the real process vector triggers the fault diagnosis process. But there is need to set a suitable threshold for a significant change in the process vector, unaffected by noisy readings or any other insignificant errors. This threshold is also dependent on the sensitivity of each sensor used for measuring the process vector. The tuning of this detection threshold was tested and chosen based on the sensitivity of the sensors on the physical system.

\subsection{Fault Localization}
In order to do detect which process parameter changed, we address the problem as a machine learning (ML) classification task. Therefore, we create a training database that samples the working region of the control vector or in some cases uses the specific operational control vector at the moment when the fault detection algorithm is triggered. The values of the resulting process vector, $Y^{i}_{s}$, and the index of the changed parameter, $n$, are stored in the database. The job of the classifier is to fit the process vector and the changed parameter into a function, $\psi$, where for the $ith$ term this becomes:
\begin{equation}
\psi(U,Y^i_{s}) =  n
\end{equation}
Therefore, with a new measured value, $Y_{m}$, after the fault detection system is triggered, the classifier predicts the perturbed parameter using the trained function, $\psi$, this becomes:
\begin{equation}
\psi(U,Y_{m}) =  n'
\end{equation}
The Decision Tree is the ML model used for this classification task.
\\

\subsection{Parameter Estimation}
Once the fault is accurately located, the process then estimates the new value of the component vector. In this regression task, the model finds the function, $g$, which best fits the process vector, $Y^{i}_{s}$, to the value of the changed parameter, $\theta^i_{n}$.
\begin{equation}
g(U,Y^i_{s},n) =  \theta^i_{n}
\end{equation}
Therefore, with a new measured value, $Y_{m}$, after the fault localization is achieved, the regression model estimates the new value of the component vector using the trained function, $g$, this becomes:
\begin{equation}
g(U,Y_{m},n) =  \theta_{n}'
\end{equation}

The Estimation of the new perturbed value is carried out by a regression model, the Support Vector Regression (SVR) model.

\subsection{Parameter Validation}
As the FDD process is triggered by a set threshold, the success of the process is also determined by converging to a set threshold in comparison with the real system. This threshold is either the same as or less than that with the detection alarm. The new estimated component vector, $\theta_e$, is simulated on the 1D model with the control vector, $U$, to get the estimated process vector, $Y_e$, and then compared with the system process vector, $Y_s$. The tuning of this threshold was tested and chosen in order to be less than the sensitivity of the sensors. The iteration of the loop was also selected as 3 but will be optimized in future works.

\begin{align}\label{eqn:validation}
\varepsilon^k &= |y_{s}^{k} - y_{m}^{k}|\\
\varepsilon^k &< \varepsilon^k_v
\end{align}

\begin{where}
  \item $k$ is the index of the vector from 1 to 5 for the process vector.
  \item $\varepsilon^k$ is the k-th error value in the error vector.
  \item $y_{s}^{k}$ is the k-th value of the measured system process vector.
  \item $y_{m}^{k}$ is the k-th value of the simulated 1D model process vector.
  \item $\varepsilon^k_v$ is the k-th value of the set validation threshold.
\end{where}

\begin{figure}[htbp]
\centerline{\includegraphics[width=1\columnwidth]{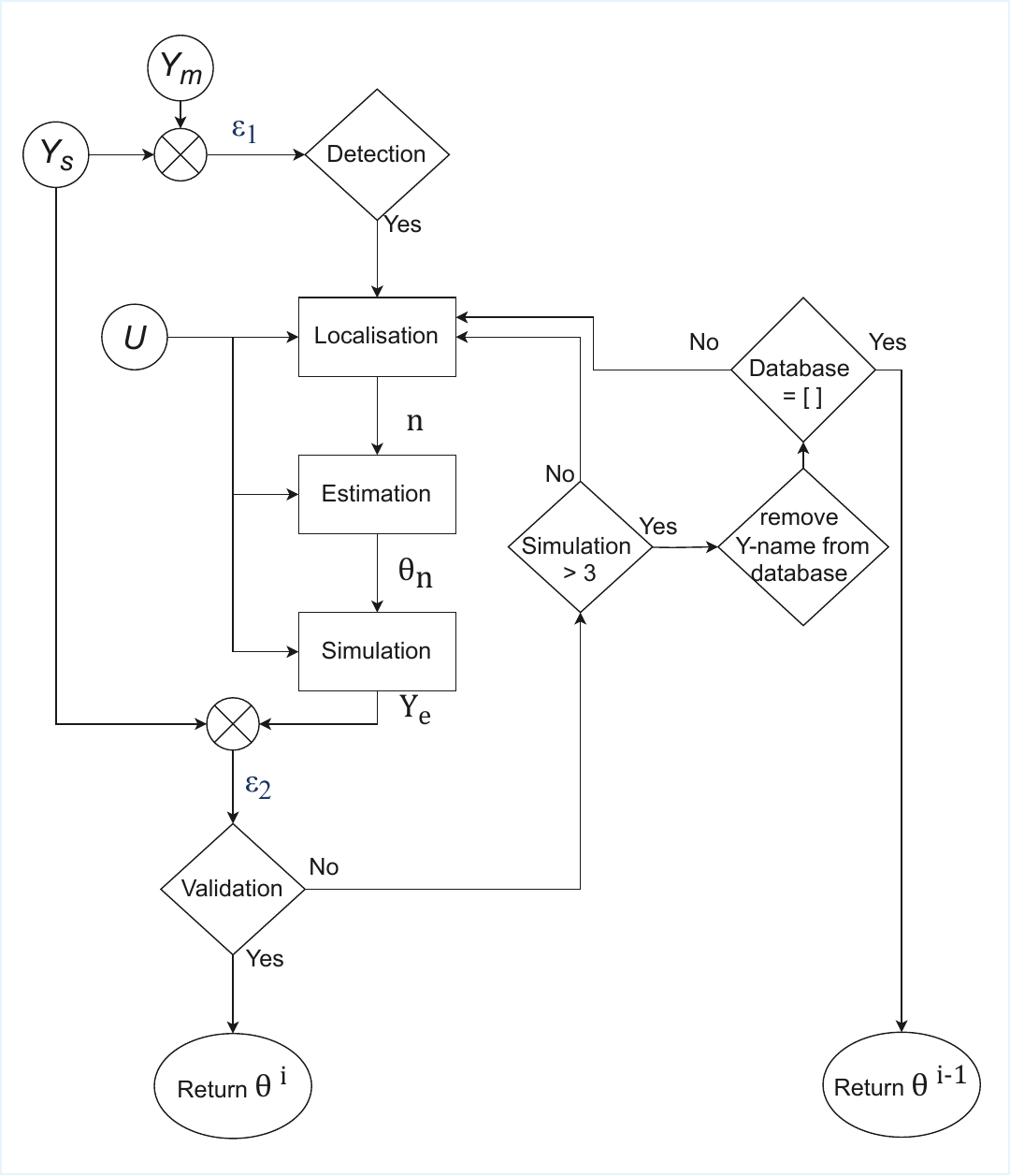}}
\caption{The FDD Algorithm}
\label{fig:FUM_JNEM}
\end{figure}

The main steps of the resulting FDD algorithm are summarized in Figure \ref{fig:FUM_JNEM}. 

\section{Experiments}
\begin{figure}[h]
    \begin{center}
        \includegraphics[width=\columnwidth]{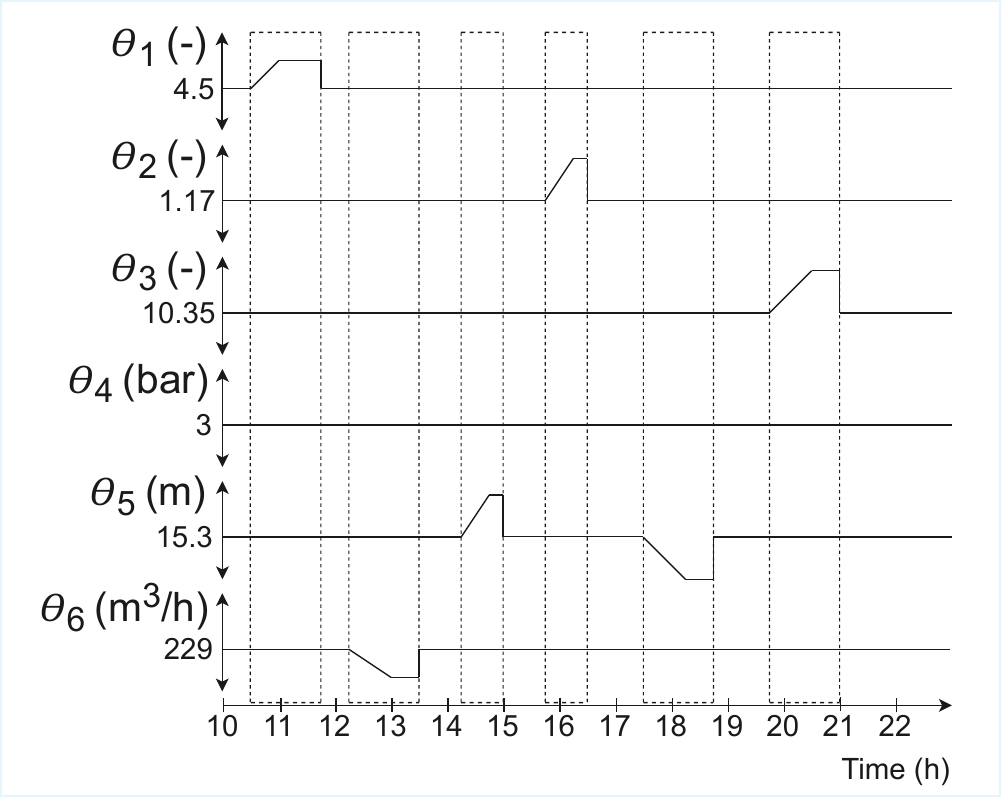}
        \caption{Example of a single perturbation scenario}
        \label{fig:Sing_pert}
    \end{center}
\end{figure}

In this paper the single perturbation occurrences have been highlighted to test the methodology. Figure \ref{fig:Sing_pert} represents graphically this assumption, with the component vector returned each time to the nominal value after the FDD process. In this section, the results of some of the numerical simulations are presented in order to validate the different steps of the methodology.

\subsection{Database}
For the moment, machine learning models are only built from the 1D model. Depending on the perturbation case, one or more of the component vector parameters are changed from the nominal while the others are left at the nominal values. In the case of single perturbation, which is tested extensively as the first stage of the doctoral work, the database is created by changing only one parameter in the component vector. 

These values are set as input for the 1D Flomaster model and the resulting process vectors and index of the changed parameter are concatenated respectively. There are three saved databases used so far in the doctoral studies, each of different sizes, differing sample rates in the control vector working space, and also differing sample rates for the changed values of each parameter. Each data-point simulated results in about 2 seconds of simulation and each database constructed offline. The database used in the following section required about 170 hours CPU for 305760 points simulated. The largest database with a smaller sampling rate is used as it performed better in all cases. It shows the best compromise between accuracy and size. In the future, work will be done to improve on the choice of database (size, regular or non-regular sampling, etc.). It would also be interesting in the future to use data from both the 1D model and the bench as the learning base for machine learning models.

\subsection{Localization}
The classification model has an accuracy of 95.14$\%$. Tables \ref{tab1} and \ref{tab2} show the accuracy of the fault localization. $\theta_5$ and $\theta_6$ are expressed together because they have the same impact on all the measurements in the hydraulic loop, and there is coupling between these two parameters. This was studied with sensitivity analysis carried out on the database, they are only distinguishable at certain values of $u_2$. It is not feasible to only implement the system at such values, hence other methods to differentiate between both are being studied.

\begin{table}[htbp]
\caption{\label{tab1} Confusion Matrix of the Fault Localization}
\begin{center}
\begin{tabular}{|c|c|c|c|c|c|}
\hline
\textbf{}&\multicolumn{5}{|c|}{\textbf{Truth}} \\
\cline{2-6} 
\textbf{Classification} & \textbf{\textit{$\theta_1$}}& \textbf{\textit{$\theta_2$}} & \textbf{\textit{$\theta_3$}}  & \textbf{\textit{$\theta_4$}}  & \textbf{\textit{$\theta_5$ $\&$ $\theta_6$}} \\
\hline
\textbf{\textit{$\theta_1$}} & \textbf{5139} & 2 & 115 & 0 & 564 \\
\hline
\textbf{\textit{$\theta_2$}} & 0 & \textbf{5702} & 141 & 0 & 157 \\
\hline
\textbf{\textit{$\theta_3$}} & 9 & 25 & \textbf{5639} & 0 & 327 \\
\hline
\textbf{\textit{$\theta_4$}} & 0 & 0 & 0 & \textbf{5699} & 0 \\
\hline
\textbf{\textit{$\theta_5$ $\&$ $\theta_6$}} & 17 & 0 & 379 & 0 & \textbf{11304} \\
\hline
\end{tabular}
\end{center}
\end{table}
\begin{table}[htbp]
\caption{\label{tab2} Accuracy Matrix of the Fault Localization}
\begin{center}
\begin{tabular}{|c|c|c|c|c|c|}
\hline
\textbf{}&\multicolumn{5}{|c|}{\textbf{Truth}} \\
\cline{2-6} 
\textbf{Classification} & \textbf{\textit{$\theta_1$}}& \textbf{\textit{$\theta_2$}} & \textbf{\textit{$\theta_3$}}  & \textbf{\textit{$\theta_4$}}  & \textbf{\textit{$\theta_5$ $\&$ $\theta_6$}} \\
\hline
\textbf{\textit{$\theta_1$}} & \textbf{88.3$\%$} & 0.03$\%$ & 1.98$\%$ & 0.00$\%$ & 9.69$\%$  \\
\hline
\textbf{\textit{$\theta_2$}} & 0.00$\%$ & \textbf{95.03$\%$} & 2.35$\%$ & 0.00$\%$ & 2.62$\%$  \\
\hline
\textbf{\textit{$\theta_3$}} & 0.15$\%$ & 0.42$\%$ & \textbf{93.98$\%$} & 0.00$\%$ & 5.45$\%$  \\
\hline
\textbf{\textit{$\theta_4$}} & 0.00$\%$ & 0.00$\%$ & 0.00$\%$ & \textbf{100.00$\%$} & 0.00$\%$  \\
\hline
\textbf{\textit{$\theta_5$ $\&$ $\theta_6$}} & 0.15$\%$ & 0.00$\%$ & 3.24$\%$ & 0.00$\%$ & \textbf{96.62$\%$}  \\
\hline
\end{tabular}
\end{center}
\end{table}


\begin{figure*}
    \centering
        \includegraphics[width=1.1\textwidth]{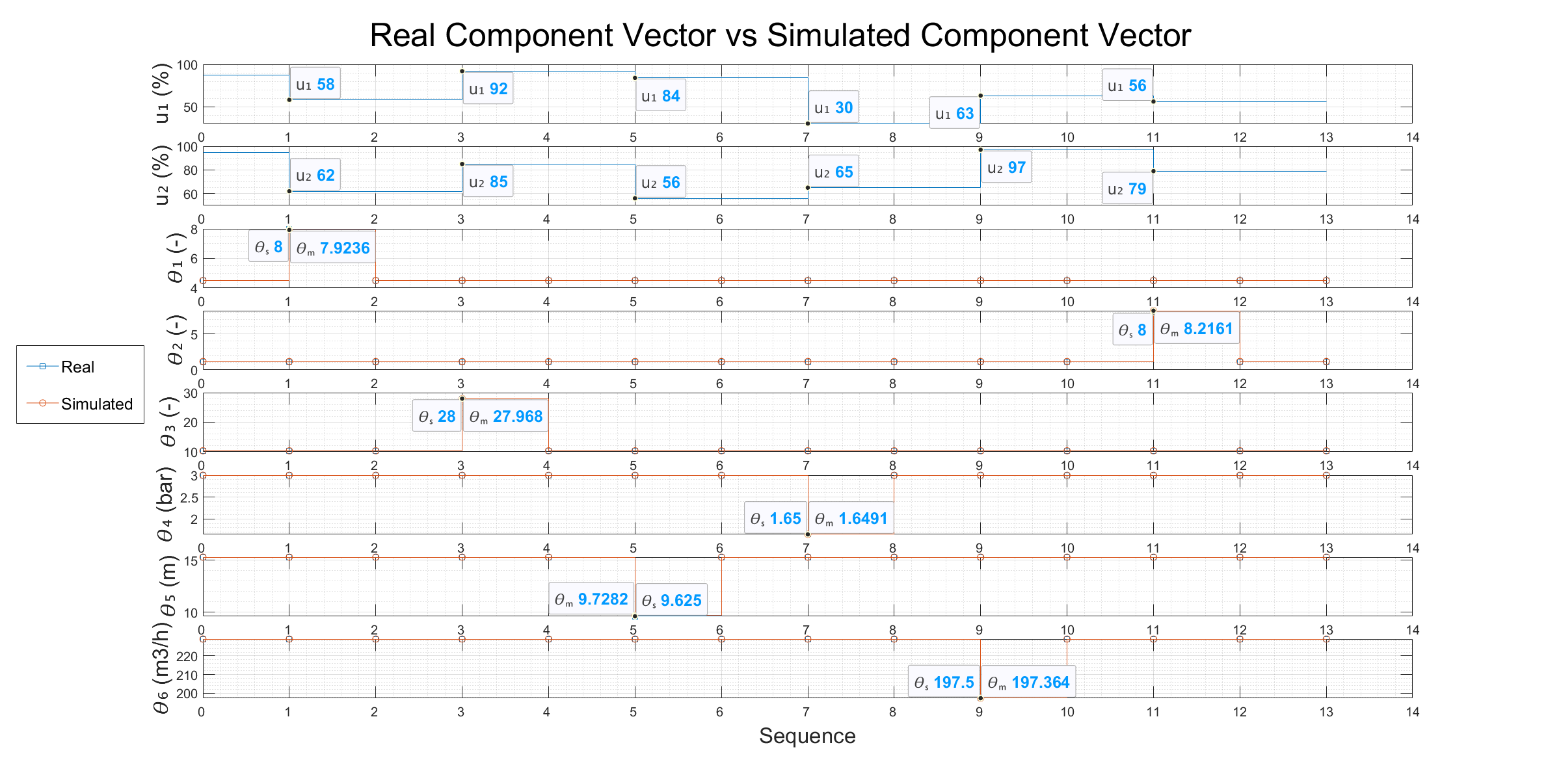}
        \caption{Evolution of single fault occurrences}
        \label{fig:Comp}
\end{figure*}

\begin{figure}[h]
    \begin{center}
        \includegraphics[width=0.5\textwidth]{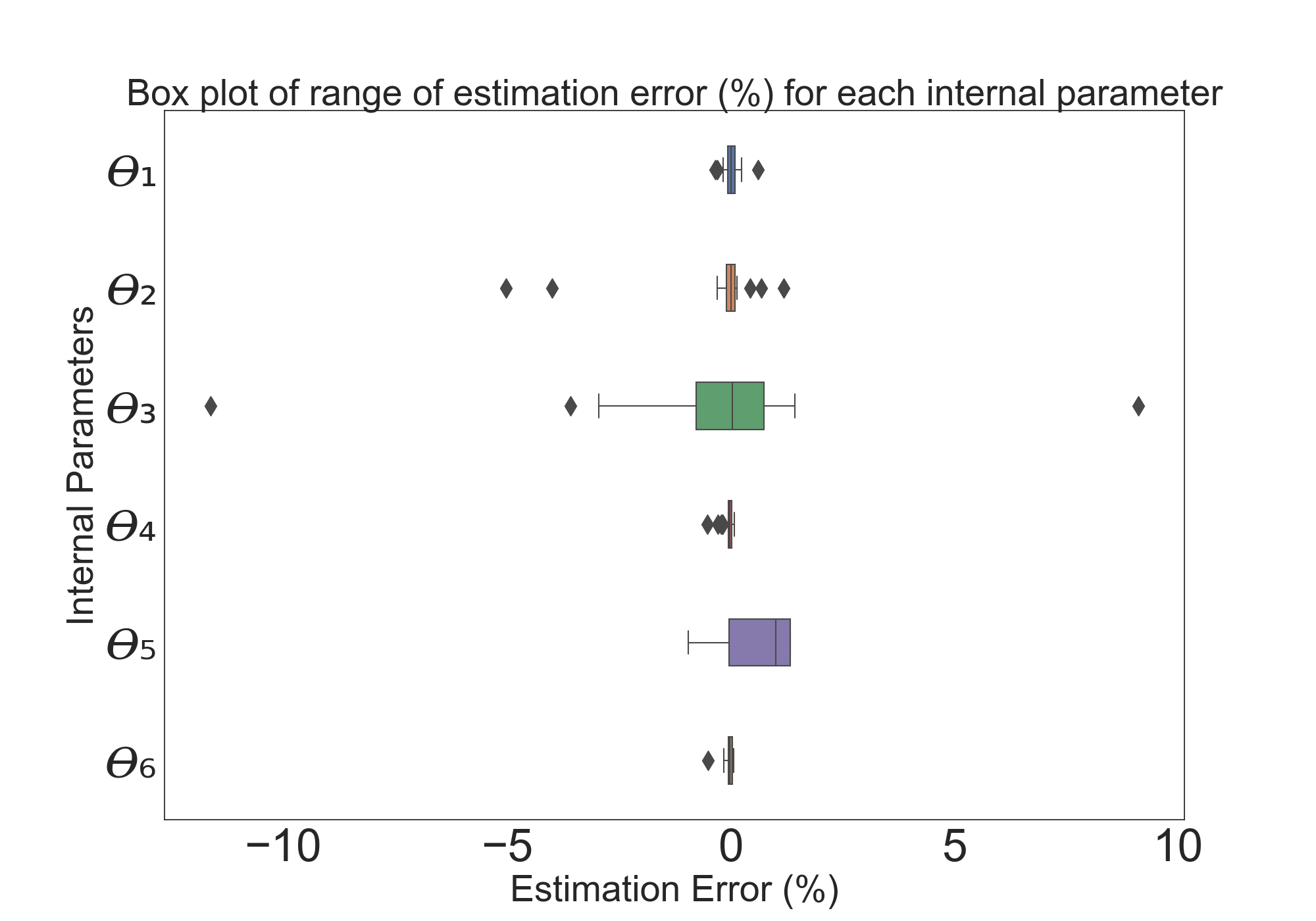}
        \caption{Estimation Error of the SVR model}
        \label{fig:EE_SVR}
    \end{center}
\end{figure}

\subsection{Estimation}
There is dependency in the proposed algorithm; if the localization is wrong, the estimation will definitely be wrong. In order to show the accuracy of the estimation model, in Figure \ref{fig:EE_SVR}, percentage estimation error for each parameter is shown in the cases where the localization is correct.

\subsection{Detection $\&$ Validation}
Several simulations were done on the overall algorithm. Figure \ref{fig:Comp} represents one of the test scenarios. The algorithm is initiated when the detection threshold is exceeded and terminates when returned within the validation threshold. The values used for the detection and validation threshold are set as:

\begin{equation}\label{eqn:i_validation}
\vect{\varepsilon}_d = \begin{bmatrix} 0.02 bar\\ 0.02 bar\\ 0.02 bar \\ 0.02 bar \\ 1 m^3/h \end{bmatrix} , \vect{\varepsilon}_v = \begin{bmatrix}
0.01 bar\\ 0.01 bar\\ 0.01 bar \\ 0.01 bar \\ 1 m^3/h     
\end{bmatrix}
\end{equation}




\section{Conclusions}
In this paper, the fault detection and diagnosis algorithm was presented which is the backbone of the physics-based black-box digital twin of a closed hydraulic loop. The algorithm is defined in four key steps: detection with a tested threshold, localization with a decision tree, estimation with SVR and 
validation with another tested threshold. This proposed algorithm has been validated in the single perturbation case with high accuracy. In future work, other perturbation cases will be modeled, following the life-cycle of the system. These will include multiple perturbations that occur either simultaneously or sequentially. Future developments include optimization of the detection, validation thresholds and database size. For the moment these are manually tuned but other adaptive methods can be explored. It is also possible to augment the model-based dataset with the data gotten from the real system,  starting with nominal points. It also concerns robustness to noise analysis before implementing the algorithm on the physical twin. The noisy analysis will be designed by modifying the test data using a random Gaussian additive noise of a set percentage. The working model will be integrated the industrial PC of the physical twin. The present model could be used in PID control even though it is not reactive and could be used in its present state for a slow not so dynamic system. It can be used in the maintenance plan for predictive maintenance. The challenges of physical application in complex systems include trying to be generic enough to be applied in other systems. Therefore for complex systems, it would require sensitivity studies and creating subsystems for model reduction and compartmental approach.
 

\end{document}